\documentclass[10pt,twocolumn,letterpaper]{article}

\usepackage[pagenumbers]{wacv} 

\usepackage{graphicx}
\usepackage{amsmath}
\usepackage{amssymb}
\usepackage{booktabs}
\usepackage{commath}
\usepackage{xcolor}
\usepackage{mathtools}
\usepackage{amsfonts}
\usepackage{multirow}
\usepackage{adjustbox}

\usepackage[pagebackref,breaklinks,colorlinks]{hyperref}

\usepackage[capitalize]{cleveref}
\crefname{section}{Sec.}{Secs.}
\Crefname{section}{Section}{Sections}
\Crefname{table}{Table}{Tables}
\crefname{table}{Tab.}{Tabs.}

\def\wacvPaperID{2158} 
\def\confName{WACV}
\def\confYear{2024}

\begin{document}

\title{Transparent Anomaly Detection via Concept-based Explanations}

\author{Laya Rafiee Sevyeri$^{*1}$\\
{\tt\small laya.rafiee@gmail.com}
\and
Ivaxi Sheth$^{*2}$\\
{\tt\small ivaxi.sheth@cispa.de}
\and
Farhood Farahnak$^{*1}$\\
{\tt\small farhood.farahnak@gmail.com}
\and
Samira Ebrahimi Kahou$^{3}$\\
{\tt\small samira.ebrahimi.kahou@gmail.com}
\and
Shirin Abbasinejad Enger$^{1}$\\
{\tt\small shirin.enger@mcgill.ca}\\
$^{1}$ Department of Oncology, McGill University, Montréal, Canada \\
$^{2}$ CISPA Helmholtz Center for Information Security \\
$^{3}$ École de Technologie Supérieure, Mila-Quebec AI, CIFAR AI Chair \\
}

\maketitle

\begin{abstract}
Advancements in deep learning techniques have given a boost to the performance of anomaly detection. However, real-world and safety-critical applications demand a level of transparency and reasoning beyond accuracy. The task of anomaly detection (AD) focuses on finding whether a given sample follows the learned distribution. Existing methods lack the ability to reason with clear explanations for their outcomes. Hence to overcome this challenge, we propose Transparent \textbf{A}nomaly Detection \textbf{C}oncept \textbf{E}xplanations (ACE). ACE is able to provide human interpretable explanations in the form of concepts along with anomaly prediction. To the best of our knowledge, this is the first paper that proposes interpretable by-design anomaly detection. In addition to promoting transparency in AD, it allows for effective human-model interaction. Our proposed model shows either higher or comparable results to black-box uninterpretable models. We validate the performance of ACE across three realistic datasets - bird classification on CUB-200-2011, challenging histopathology slide image classification on TIL-WSI-TCGA, and gender classification on CelebA. We further demonstrate that our concept learning paradigm can be seamlessly integrated with other classification-based AD methods.

\end{abstract}
\section{Introduction}\label{sec:intro}
\def\thefootnote{*}\footnotetext{These authors contributed equally to this work.}
In recent years, deep learning models have achieved remarkable advancements, often demonstrating performance levels on par with human capabilities in a wide array of tasks like image segmentation~\cite{kirillov2023sam}, image generation~\cite{nichol2021glide}, and text generation~\cite{lewis2020bart,raffel2020t5}. As a result, these models have been increasingly applied in diverse real-world contexts. However, it's crucial to recognize that despite their achievements, instances of model failure have surfaced in practical scenarios, raising concerns about their reliability for safety-critical applications. One contributing factor to these failures is rooted in the assumption that the distribution used for testing deep learning models matches their training distribution. However, this assumption does not hold in many realistic tasks limiting their application. Hence it is important for the model to be able to differentiate between the different distributions. The ability to identify and adapt to out-of-distribution (also known as anomalous) instances is vital for ensuring that the model's predictions remain reliable and trustworthy in novel and diverse scenarios which can be aided by anomaly detection. Anomaly detection models aims to differentiate between data points that follow a certain distribution and those that deviate from it.

In many critical domains such as healthcare and finance, it is not only crucial to identify anomalies but also to provide meaningful explanations for the detected anomalies. This is particularly important to enhance user trust, facilitate decision-making, and ensure the accountability of deep learning models.
\begin{figure}
    \centering
    \includegraphics[scale=0.15]{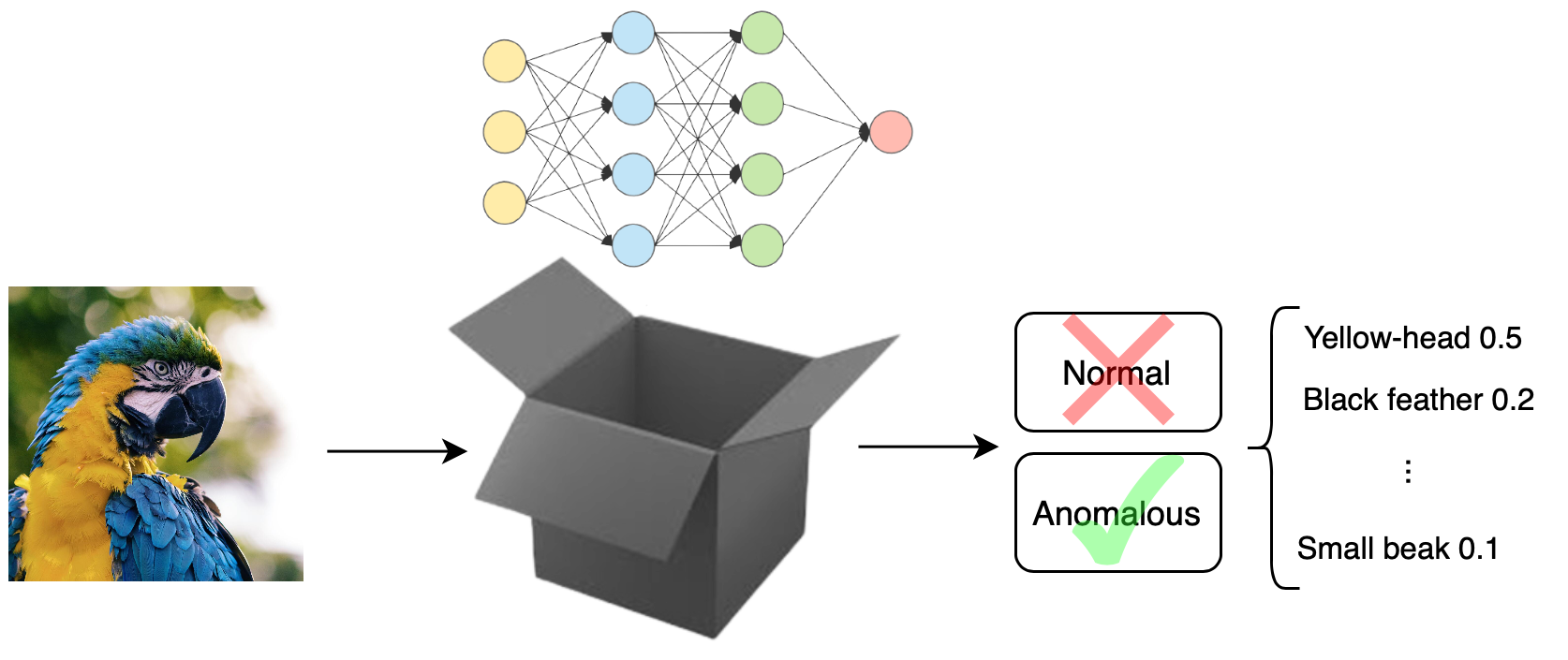}
    \caption{ACE; Anomaly detection with ACE on the CUB dataset. Corresponding concepts provide detailed insights and explanations into model behavior.}
    \label{fig:ace}
\end{figure}

The task of anomaly detection has been of great interest in the research community and has developed strong methods to detect distributional shifts~\cite{tax2004SVDD,schlegl2017AnoGAN,Bergman2020GOAD,reiss2023msc}. Most of these works have focused on improving the algorithms to accurately discriminate between in-distribution (also known as normal) data and out-of-distribution (anomalous) instances~\cite{scholkopf1999OC_SVM,zong2018gaussian_mixture,Bergman2020GOAD,Golan2018GEOM,reiss2023msc}. 
However, despite the progress in deep learning, a critical aspect - explainability - has been overlooked for anomaly detection methods. An ``easy''and obvious method could be to use the interpretability methods for each in-domain and out-domain data separately and then visualize the difference~\cite{sundararajan2017axiomatic}. While this has been somewhat explored in literature, it could be argued that pixel-level activations may not necessarily be the most effective methods for human understanding~\cite{adebayo2020debugging}.

Since the emergence of research focused on anomaly detection, there have been different subgroups of problem statements introduced~\cite{chandola2009anomaly}. Through our work, we aim to bridge the research gap between human-level explainability and anomaly detection using human-understandable concepts. To the best of our knowledge, this is the first work that introduces an inherently interpretable by-design model for anomaly detection. Concept-based explanations provide an interpretable linkage between the model's decisions and the high-level concepts learned during training. By conditioning the detection of anomalous (out-of-distribution) instances with specific concepts, we not only offer insights into the model's decision-making process but also empower domain experts to validate the model's conclusions. 

One popular approach for unsupervised anomaly detection is using self-supervised learning to train an auxiliary task (e.g. discriminating between different transformations) to learn better representations for normal training data and consequently improving the model's discriminativeness~\cite{Bergman2020GOAD,reiss2023msc,tack2020csi,Golan2018GEOM}. 
In this work, we propose Transparent Anomaly Detection Concept Explanations (ACE) that provide explanations for transformation-based anomaly detection models (see Fig.~\ref{fig:ace}). In addition to providing explanations, our method can allow a domain expert to interact with the model using concept-based explanations. This interaction allows the supervisor to correct the concepts if they disagree with the model's explanations which can improve the downstream representation as well. We conduct extensive experiments on diverse datasets to showcase the effectiveness of our approach. Our results demonstrate that our concept-based explanation framework not only enhances the interpretability of anomaly detection models but also maintains competitive detection performance.

\section{Related Work}
\label{sec:related_work}
In this section, we will review some of the previous studies related to out-of-distribution detection and concept learning.

\paragraph{Anomaly Detection}
Anomaly detection (AD) or in general out-of-distribution (OOD) detection approaches can be grouped according to the following paradigms.
\textbf{Distributional-based approaches} try to build a probabilistic model on the distribution of normal data. They rely on the idea that the anomalous samples would act differently than the normal data. They expect that the anomalous samples receive a lower likelihood under the probabilistic model than the normal samples. The difference in these models is in the choice of their probabilistic model and their feature representation approach. Gaussian mixture models~\cite{parzen1962estimation}, which only work if the data can be modeled with the probabilistic assumptions of the model, and kernel density estimation (KDE)~\cite{latecki2007KDE} methods are among traditional methods. Some recent approaches use deep learning to represent the features~\cite{zhou2017anomaly, yang2017towards}. To alleviate the limitation that the probabilistic assumption imposes, recent studies suggested learning a probabilistic model on the features extracted by the deep models~\cite{zong2018gaussian_mixture}.
\textbf{Classification-based approaches} One-Class SVM (OC-SVM)\cite{scholkopf1999OC_SVM} and support vector data description (SVDD)~\cite{tax2004SVDD} are among the first works in this category. They used the idea of separating the normal data from the anomalous data based on their feature spaces. In the long history of the studies of this paradigm, different approaches from kernel methods to deep learning approaches such as Deep-SVDD~\cite{ruff2018deep_SVDD} have been used. However, these approaches may suffer from the insufficient and biased representations the feature learning methods can provide. One remedy for this issue is using self-supervised learning methods. Various surrogate tasks such as image colorization~\cite{zhang2016colorful}, video frame prediction~\cite{Mathieu2016video_fram_prediction}, and localization~\cite{yang2021insloc} are among those that provide high-quality feature representations for downstream tasks. In 2018, Golan \etal~\cite{Golan2018GEOM} proposed geometric transformation classification (GEOM) to predict different geometric image transformations as their surrogate task for anomaly detection. Following that, Bergman \etal~\cite{Bergman2020GOAD} introduced GOAD, a unified method on one-class classification and transformation-based classification methods. Sohn \etal~\cite{sohn2021learning} presented a two-stage framework with a self-supervised model to obtain high-level data representations as the first stage, followed by a one-class classifier, such as OC-SVM or KDE, on top of the representations of the first stage. Whereas CSI~\cite{tack2020csi} changed the conventional contrastive learning setting for anomaly detection by contrasting each example by distributionally-shifted augmentations of itself. MSC~\cite{reiss2023msc} recently proposed a new contrastive objective to use transformed representations pretrained on an external dataset for anomaly detection.
\textbf{Reconstruction-based approaches} instead of relying on the lower likelihood of the distributional-based methods, these approaches rely on the idea that normal samples should receive smaller reconstruction loss rather than anomalous samples. Different loss and reconstruction basis functions vary in each of these approaches. K-means is used as an early basis reconstruction function~\cite{jianliang2009k_meansAD} while \cite{an2015variational} proposed using deep neural networks as the basis functions. In the class of deep neural networks, generative models such as GANs~\cite{schlegl2017AnoGAN} and autoencoder~\cite{zhou2017anomaly} are used to learn the reconstruction basis functions. Following the presentation of AnoGAN~\cite{schlegl2017AnoGAN} as the first anomaly detection model based on GAN, several other studies used similar ideas with modifications on their basis functions and losses~\cite{zenati2018ALAD,deecke2018ADGAN,zenati2018efficient,rafiee2020unsupervised} to increase the performance of anomaly detection models based on GANs. One of the major issues in using generative models, especially GANs as the reconstruction basis function, is their difficulty in recovering the entire data distribution (aka mode-collapse in GANs),  leading to lower performance in comparison with classification-based approaches. \cite{sevyeri2022adcgan} combined adversarial training with contrastive learning to mitigate the challenges of reconstruction-based approaches.

\paragraph{Interpretable Anomaly Detection}
While ample research has been conducted in the field of anomaly detection (AD), only a limited number of studies have focused on developing interpretable AD models. Carletti \etal~\cite{carletti2023interpretable} employed feature importance to improve the explaianability to the Isolation Forest. While an interpretable and explainable anomaly detector has been rarely explored, few studies improved explainablity using concepts to address distribution shifts ~\cite{kim2018interpretability,adebayo2020debugging}.
More specifically on out-of-distribution detection,~\cite{xu2023interpretable} proposed the idea of introducing a new confidence measure based on PARTICUL~\cite{xu2022particul}, an existing pattern identification model into OOD detection. Seras \etal~\cite{seras2022novel} relied on the attribution maps in Spiking Neural Networks (SNNs)~\cite{ghosh2009snn} to explain the reason behind the prediction of the model. Szymanowicz~\etal~\cite{szymanowicz2022discrete} proposed an explainable OOD detection based on saliency maps in video. On the other hand,~\cite{doshi2023towards} proposed a dual-monitoring approach involving global and local elements to construct a scene graph that observes interactions among different objects within a scene. In a rather limited attempt, Cho \etal~\cite{cho2023training} introduced a new semi-supervised explainable OOD detection model for medical image segmentation. This was achieved by training an auxiliary prototypical network based on outlier exposure~\cite{hendrycks2018deep}.

\paragraph{Concept-based Explainability}
Concept Bottleneck Models (CBMs)~\cite{koh2020cbm} introduced explainability by adding predefined human understandable features in the neural network. 
Despite the popularity of CBMs, \cite{mahinpei2021promises} pointed out that their concept representations may lead to information leakage, thus diminishing their performance. Since its introduction, there have been various works that build upon CBMs to overcome their shortcomings~\cite{espinosa2022concept, havasi2022addressing, sheth2023overcoming, kim2023probabilistic}. Havasi \etal\cite{havasi2022addressing} showed that CBM performance is highly dependent on the set of concepts and the expressiveness of the concept predictor and modified CBMs using autoregressive models and disentangled representations. \cite{espinosa2022concept} improved the performance of concept-based models by introducing high-dimensional concept embedding. Sheth \etal~\cite{sheth2023overcoming} proposed a multi-task concept learning model for medical imaging applications.
Beyond explainability, CBMs also allow human intervention to refine the concepts during inference. These interventions were further studied by \cite{shin2022a, chauhan2023interactive, sheth2022learning, zarlenga2023learning}.

\begin{figure*}
    \centering
    \includegraphics[scale=0.15]{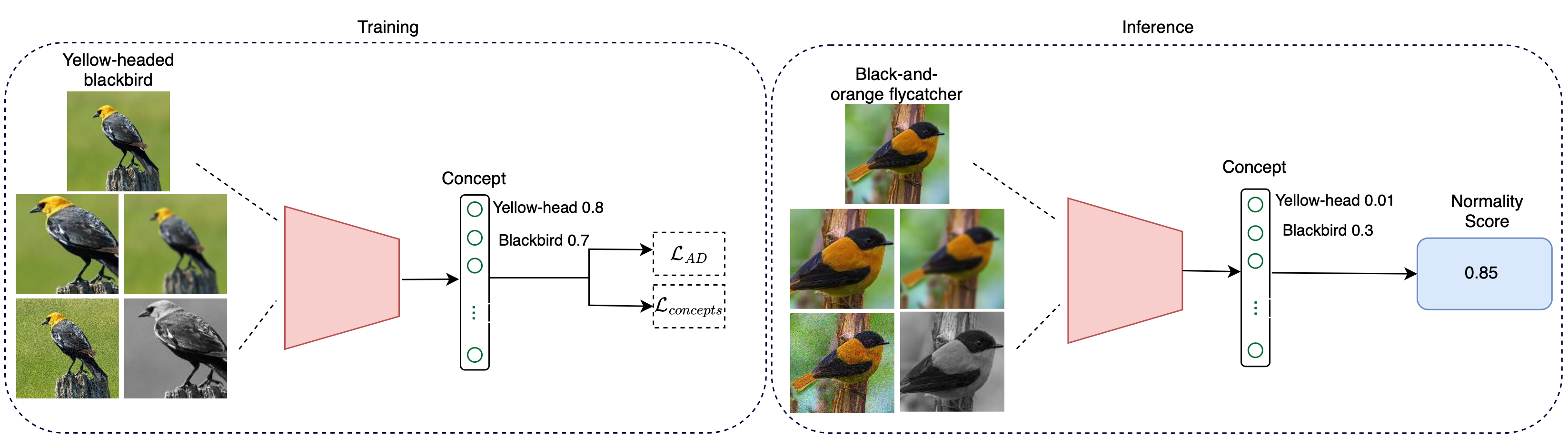}
    \caption{ACE; Training and inference of ACE on two examples from the CUB dataset with their corresponding transformed images. Since the example in the inference is from different classes as the normal training example, it receives a higher normality score indicating anomalous. }
    \label{fig:ace_fullmodel}
\end{figure*}
\section{Methodology}
\subsection{Background}
In an unsupervised anomaly detection, models only have access to the normal training data. Once the unsupervised anomaly model is trained, the representations obtained from it will be used to separate normal and anomalous samples. One-class (OC) classification models rely on a normality score defined to identify anomalous samples. While these models achieved state-of-the-art performance in detecting anomalies, they lack explanatory capabilities for their predictions. Therefore, the incorporation of concepts into the detection process introduces a more transparent anomaly detection model.

\subsection{ACE}
In this section, we deconstruct the Transparent \textbf{A}nomaly Detection \textbf{C}oncept \textbf{E}xplanations (ACE) model into two distinct modules: anomaly detection and concept-based explanation. We further explain each module in the following section.

\paragraph{Formal Definition}
In our AD task, assuming that all data lies in $R^L$ where $L$ defines the data distribution, we defines normal (in-distribution) data as a subspace $X \subset R^L$ which includes $x \in X$. Therefore, given an unsupervised setting, the training set $D_{train} = \{x_1, x_2, ..., x_k\sim P_{ind}\}$ contains normal samples drawn from $P_{ind}$, in-distribution. To evaluate the model, we use a test set $D_{test} = \{\bar{x_1}, \bar{x_2},..., \bar{x_n}\sim P_{ind}~\cup~P_{ood}\}$ including both normal and anomalous samples drawn from in and out-of-distribution ($P_{ood}$) respectively. Our model belongs to the category of unsupervised approaches as it does not see any out-of-distribution sample during training.

\paragraph{Concept Explanations}
In the concept-based model setup, each dataset is augmented with its corresponding auxiliary human interpretable concepts. For ACE, we reformulate the training dataset used for anomaly detection to incorporate concepts. As a result, the training dataset is redefined as $D_{train} = \{(x_1, c_1), (x_2, c_2), ..., (x_k, c_k)\sim P_{ind}\}$, where $c \in \mathbb{R}^0$. Each concept vectors is of the length $N$ denoting the number of human-interpretable concepts we train with.
Our concept representation is binary in nature i.e. the model predicts whether a concept is present in the image or not. While traditional concept based models were developed for supervised classification, we modify the objective for anomaly detection. The first step is to extract the concepts from the image, we define a concept model, an encoder $\mathcal{G}_{X \to C}$, to map each image $x \in X$ into the concept $c$. In the concept encoder, most of the parameters are shared apart from the very shallow concept prediction layer. For brevity, we continue to denote the concept encoder with concept prediction layers as $\mathcal{G}_{X \to C}$.
To train the concepts in ACE, we use binary cross entropy loss function across each concept:
\begin{equation}\label{eq:loss_concept}
    \mathcal{L}_{concepts}(c, \hat{c}) = \sum_{c_i}^{c_N} -c_i \textup{log}(\hat{c}_i) - (1-c_i) \textup{log} (1-\hat{c}_i)
\end{equation}
\noindent where $\hat{c}$ is the predicted concept.

\paragraph{Anomaly Detection}
In this work, we follow the transformation-based classification methods for anomaly detection~\cite{Bergman2020GOAD,Golan2018GEOM,tack2020csi,reiss2023msc}. Given an input tuple $(x, c) \in D_{train}$ and a transformation function $T(.)$ with $M=\{t_1, t_2, ..., t_m\}$ different transformations, we apply all the transformations to each normal image $x \in X$. Hence, for each input image, the original input image $x$ and its corresponding transformed images $x^{\prime} \in X^{\prime}_m$ will be mapped to their corresponding feature representations using the encoder $\mathcal{G}_{X \to C}$. 

Each transformation $t_m \in M$ forms a cluster with a centroid $s_m$ defining the sphere. The centroid is the average of the features over all the training set for every transformation and is computed by $s_m = \frac{1}{N}\Sigma_{x \in X}\mathcal{G}(T((x, c), t_m))$ where $N$ defines the number of samples in the training set.

Following Bergman \etal~\cite{Bergman2020GOAD} and in order to have lower intra-class variation and higher inter-class variation for each cluster (feature space), we define a transformation loss using triplet loss~\cite{he2018triplet} for the training of encoder $\mathcal{G}$ as follows:
\begin{equation}\label{eq:ad_loss}
    \begin{split}
       \mathcal{L}_{AD} = \Sigma_{i}~ &max~({\norm{\mathcal{G}(T(x_i, c_i, t_m))- s_{m}}^2} + d\\
        - &min_{m^{\prime} \neq m} \norm{\mathcal{G}(T(x_i, c_i, t_m))-s_m^{\prime}}^2, 0)
    \end{split}
\end{equation}
\noindent where $d$ is a hyperparameter regularizing the distance between clusters.

The final loss for training ACE combines concept (Eq.~\ref{eq:loss_concept}) and anomaly detection (Eq.~\ref{eq:ad_loss}) losses: 
\begin{equation}\label{eq:loss_ace}
    \mathcal{L}_{ACE} = \alpha \mathcal{L}_{concepts} + \mathcal{L}_{AD}
\end{equation}

\noindent where $\alpha$ is the hyper-parameters controlling the contribution of the accurate concept learning training process.

\paragraph{Normality Score}
During the inference time, all $M$ different transformations will be applied to each sample $x \in D_{test}$ including samples from $P_{ind} \cup P_{ood}$. The probability of $x$ being identified as normal is the product of the probabilities that all transformed samples lie within their respective subspaces. Therefore, we compute the normality score as presented in Eq.~\ref{eq:normality_score}, where higher values indicates anomalous samples (see Fig~\ref{fig:ace_fullmodel} for a schematic overview of ACE).
\begin{equation}\label{eq:normality_score}
    \begin{split}
        NS (x) = - log P(x \in X) &= - \Sigma_{t_m} P(T(x, c, t_m) \in X_m) \\
        &= - \Sigma_{t_m} P(t_m | T(x, c, t_m))
    \end{split}
\end{equation}

\section{Experiments and Results}
We conduct extensive experiments across two benchmark datasets with different domains (i.e. vision and medical) to validate the performance of our model, ACE. We aim to show that our method either improves performance or has a comparable performance to its baselines with explanatory power. 

\begin{table*}[t]
\centering
\begin{adjustbox}{width=0.8\textwidth}
\begin{tabular}{lccccc}
\toprule
Datasets & Class ($k_{ind}$) & ~~ & GOAD~\cite{Bergman2020GOAD} & GOAD+ACE & Concept Accuracy\\
\midrule
& Black footed Albatross & ~~ & 71.76$\pm$~0.01 & 73.38$\pm$~0.00 & 92.47$\pm$0.34 \\
& Laysan Albatross & ~~ & 63.73$\pm$~0.02 & 65.02$\pm$~0.03  & 86.94$\pm$0.88 \\
& Sooty Albatross & ~~ & 60.02$\pm$~0.01 & 60.35$\pm$~0.02  &87.37$\pm$1.39\\
& Groove billed Ani & ~~ & 67.94$\pm$~0.01 & 69.31$\pm$~0.01  &93.11$\pm$2.20\\
CUB & Crested Auklet & ~~ & 66.31$\pm$~0.03 & 66.91$\pm$~0.03  &88.38$\pm$0.79\\
& Least Auklet & ~~ & 55.07$\pm$~0.03 & 58.84$\pm$~0.02   &84.26$\pm$2.21\\
& Parakeet Auklet & ~~ & 77.79$\pm$~0.01 & 77.85$\pm$~0.02  &93.07$\pm$0.54\\
& Rhinoceros Auklet & ~~ & 70.91$\pm$~0.01 & 70.00$\pm$~0.00 & 90.42$\pm$1.23 \\
& Brewer Blackbird & ~~ & 56.05$\pm$~0.02 & 56.44$\pm$~0.02 & 89.57$\pm$1.95\\
& Red winged Blackbird & ~~ & 77.60$\pm$~0.01 & 77.96$\pm$~0.01 & 94.04$\pm$ 1.63 \\
\midrule
& Average & & 66.72 & \textbf{67.61} & 89.96 \\

\midrule
& BLCA & ~~ & 53.40$\pm$~0.04 & 58.83$\pm$~0.00  & 90.91 $\pm$~0.98\\
& BRCA & ~~ & 54.56$\pm$~0.06 & 57.43$\pm$~0.00  & 92.68$\pm$~1.96\\
& CESC & ~~ & 51.52$\pm$~0.08 & 53.66$\pm$~0.08  & 91.63$\pm$~0.59\\
& COAD & ~~ & 40.88$\pm$~0.02 & 37.82$\pm$~0.03 &91.90 $\pm$~1.38 \\
TIL & LUAD & ~~ & 50.01$\pm$~0.01 & 52.21$\pm$~0.01 &92.14 $\pm$~1.27\\
& LUSC & ~~ & 53.75$\pm$~0.03 & 56.71$\pm$~0.05 &  91.88$\pm$~0.83\\
& PAAD & ~~ & 50.64$\pm$~0.01 & 52.92$\pm$~0.02  & 85.39$\pm$~0.48\\
& PRAD & ~~ & 56.84$\pm$~0.02 & 55.59$\pm$~0.09 & 88.47$\pm$~1.21\\
& READ & ~~ & 56.05$\pm$~0.02 & 56.44$\pm$~0.02  &89.43$\pm$~0.32\\
& SKCM & ~~ & 41.72$\pm$~0.02 & 48.92$\pm$~0.06  &91.11$\pm$~0.96\\
& STAD & ~~ & 52.28$\pm$~0.01 & 52.66$\pm$~0.00  &90.46$\pm$~0.46\\
& UCEC & ~~ & 48.96$\pm$~0.05 & 61.30$\pm$~0.02  &91.33$\pm$~0.74\\
& UVM & ~~ & 53.76$\pm$~0.06 & 62.57$\pm$~0.02  &83.45$\pm$~2.40\\

\midrule
& Average & & 51.11 & \textbf{54.39} & 83.14 \\
\midrule
& Female & ~~  & \textbf{65.75}$\pm$~0.01 & 65.28$\pm$~0.00  & 81.52$\pm$~2.39\\
CelebA & Male & ~~ & 39.20$\pm$~0.01 & \textbf{40.01}$\pm$~0.01 & 74.68$\pm$1.85\\
\midrule
& Average & ~~ & {52.47} & \textbf{52.64} & 78.08 \\
\bottomrule
\end{tabular}
\end{adjustbox}
\caption{\small{ROC-AUC ($\%$) comparison of AD models on TIL, CUB, and CelebA datasets with \textit{one-vs-all} scheme. In the \textit{one-vs-all} scheme, the class name defines $k_{ind}$. The results are averaged over five different runs. We used  $\alpha = 0.01$,  $\alpha = 1.0$, and $\alpha = 0.01$ for our experiments on TIL, CUB, and CelebA  datasets respectively. The concept accuracy is only reported for GOAD+ACE
}}\label{tab:results}
\end{table*}
\subsection{Datasets} We conduct experiments for bird classification CUB-200-2011 and cancer histopathology cancer classification TIL-WSI-TCGA data. For bird classification, we used the Caltech-UCSD Birds-200-2011 (CUB)~\cite{Welinder2010CaltechUCSDB2}. The CUB dataset has 11,788 images with $200$ classes, however, we trained our anomaly detection model on the first $10$ classes and tested on the first $20$ classes only. 
We also conducted experiments on CelebA (CelebFaces Attributes Dataset)~\cite{liu2018large}. CelebA has 202,599 face images of 10,177 celebrities, each annotated with 40 binary labels indicating facial attributes like hair color, gender, and age. In this experiment, we focus on gender classification between males and females.
For the medical dataset, we examined the Tumor-Infiltrating Lymphocytes (TIL) dataset~\cite{Saltz2018SpatialOA}, which contains histopathology images from various cancer types. TIL encompasses 13 subsets of the TCGA dataset, each representing a distinct cancer type.

In order to evaluate ACE on anomaly detection tasks, we employ \textit{one-vs-all} scheme. In this scheme, a dataset with $K$ classes will lead to $K$ different anomaly detection experiments. A given class $k_{ind}$, $1 \leq k_{ind} \leq K$, is considered as the \textit{normal} class, while $k_{ood}$ defines \textit{anomalous} class of the rest of $K-1$ classes.

\subsection{Baselines}
Our model ACE introduces concept-based explanations in the \textit{one-vs-all} category of anomaly detection. Hence we compare the performance of our model with the black-box anomaly detection model. We considered a popular transformation-based anomaly detection baseline-GOAD~\cite{Bergman2020GOAD} for all of our experiments. The aim of our model is to maintain the performance of GOAD along with providing human interpretable explanations. We further performed an ablation study by using a different backbone-MSC~\cite{reiss2023msc} instead of GOAD (see Sec.~\ref{sq:msc}).

\subsection{Experimental Setup}
We use the same hyperparameters for GOAD and GOAD+ACE for the anomaly detection task. We used $M=72$ transformations during training. For training, we employed the SGD optimizer with a learning rate of $0.01$ and a batch size of $4$. The training of GOAD and GOAD+ACE was conducted over 15 epochs for both datasets. The backbone is a WideResNet10 model~\cite{zagoruyko2016wide}. For GOAD+ACE experiments, the concept weight is $\alpha$=$0.01$.

\subsection{Results}
With extensive experiments, we demonstrate that our explainable concept-based anomaly detection approach can indeed be effectively applied to transformation-based anomaly detection models. The performance of ACE is outlined in Table~\ref{tab:results}. The Receiver Operating Characteristics (ROC) curve's Area Under the Curve (AUC) measures the classifier's performance across different threshold settings. In the context of this research, the ROC-AUC assesses the classifier's ability to differentiate between normal and anomalous samples. The results from Table~\ref{tab:results} indicate that adding concept explanations to the anomaly detection enhances both the interpretability as well as the overall performance of the model, particularly in the case of challenging TIL dataset.

To evaluate the faithfulness in concept prediction, we use concept accuracy as a metric. High concept accuracy signifies that the model is able to learn concept representation aligned to human understanding and annotation.

\subsection{Ablation Studies}
To evaluate the robustness of our model against various hyperparameters, we conducted extensive experiments.

\subsubsection{Sparse Concept Scenario}
\begin{figure}[] 
    \centering
    \includegraphics[width=0.45 \textwidth]{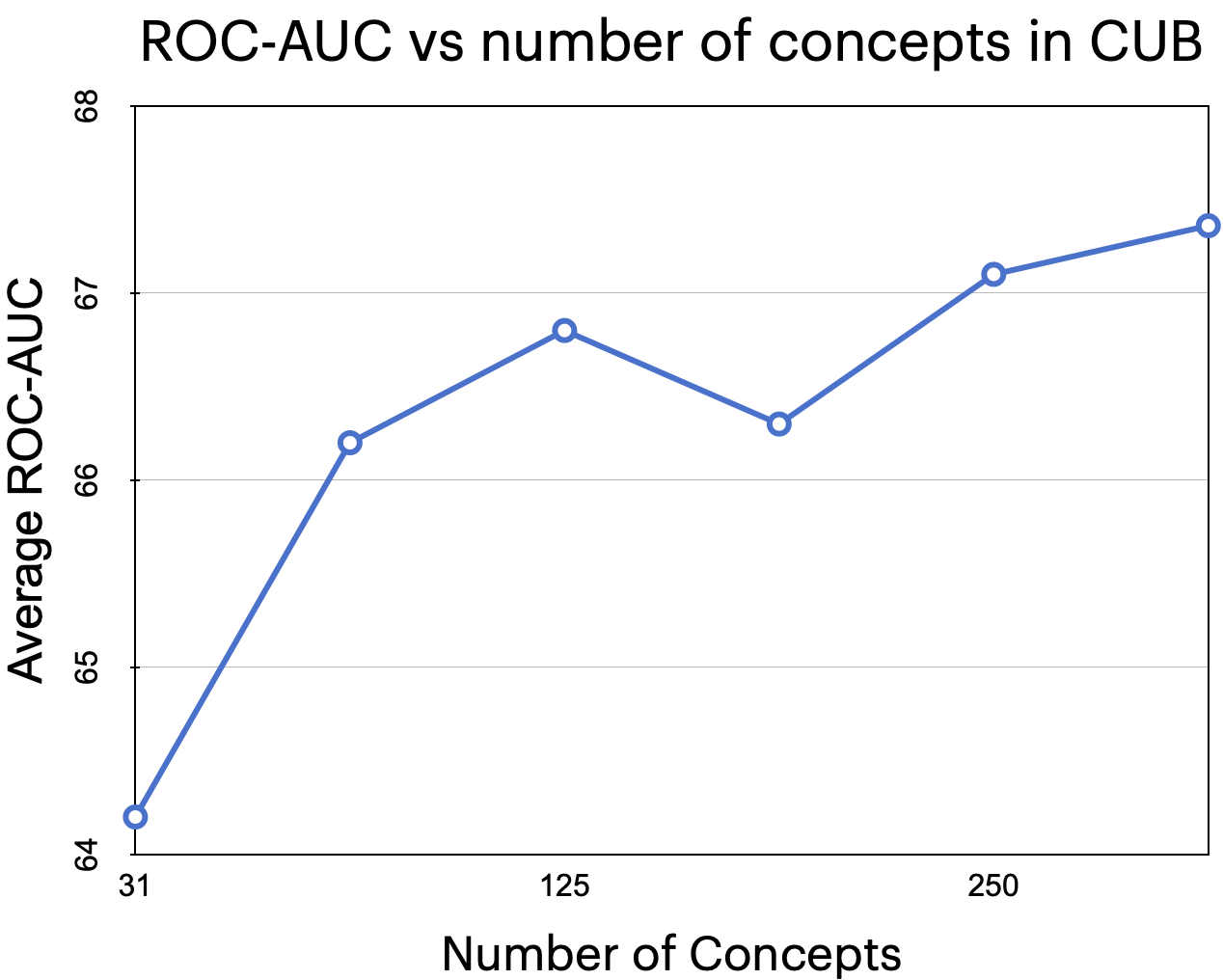} 
    \caption{Number of concepts and its impact on CUB using GOAD+ACE; In general increasing the number of concept leads to the higher ROC-AUC.}
    \label{fig:conceptvscub}
\end{figure}

\begin{figure}[] 
    \centering
    \includegraphics[width=0.45 \textwidth]{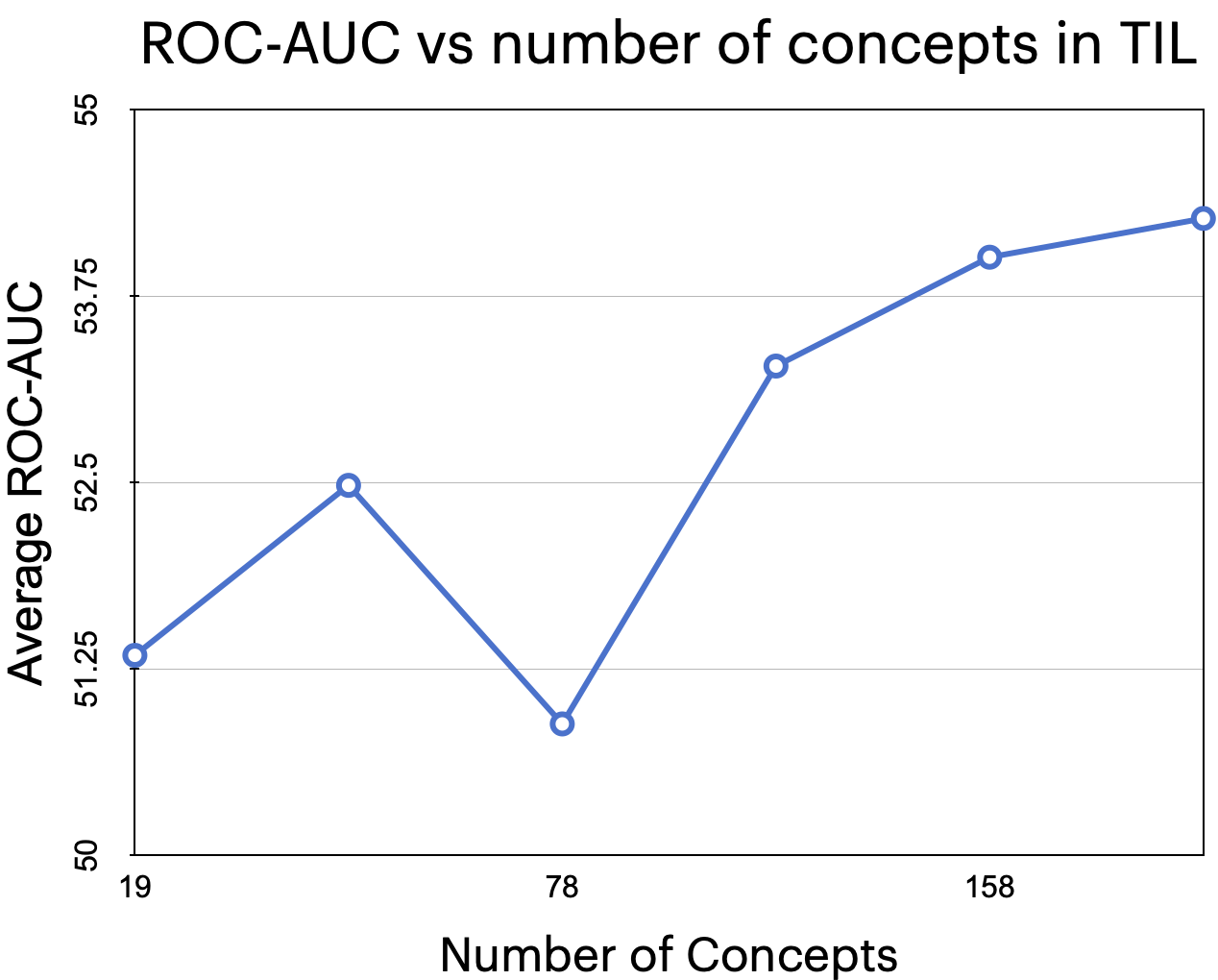} 
    \caption{Number of concepts and its impact on TIL using GOAD+ACE; Increasing the number of concept improves the performance.}
    \label{fig:conceptvstil}
\end{figure}

The concept representation for vision and medical datasets is fairly different. For medical datasets such as TIL, the concepts are fairly easily available through medical notes. However, in settings where concept data annotation is not easy to obtain, it might be difficult to gauge the optimal number of concepts to label. Although finer concept label suggests that we can obtain finer knowledge about the image. To understand the effect of concept annotation, we performed experiments by training with a part of concepts only. In the Figure~\ref{fig:conceptvscub} and Figure~\ref{fig:conceptvstil}, we averaged the AUC of each label for CUB and TIL datasets respectively. Our experiments showed that using only $10\%$ of concept annotation resulted in a lower AUC, whereas an increase in the number of concept labels led to an AUC enhancement (apart from $40\%$ concept annotation on TIL).


\subsubsection{Influence of Concept Weightage}
The hyperparameter $\alpha$ controls the weightage given to the concept learning in the final ACE loss. We conduct experiments to evaluate the sensitivity of ACE to this hyperparameter. Given the small variations in AUC for CUB dataset across $\alpha=(0.001,0.01,0.1,1.0,10.0)$, we conclude that our model is fairly robust to changes in $\alpha$ in the CUB dataset. For the medical TIL dataset, we observe a fairly similar trend. In summary, our model is robust to different concept weights while being interpretable.

\begin{figure}[] 
    \centering
    \includegraphics[width=0.45 \textwidth]{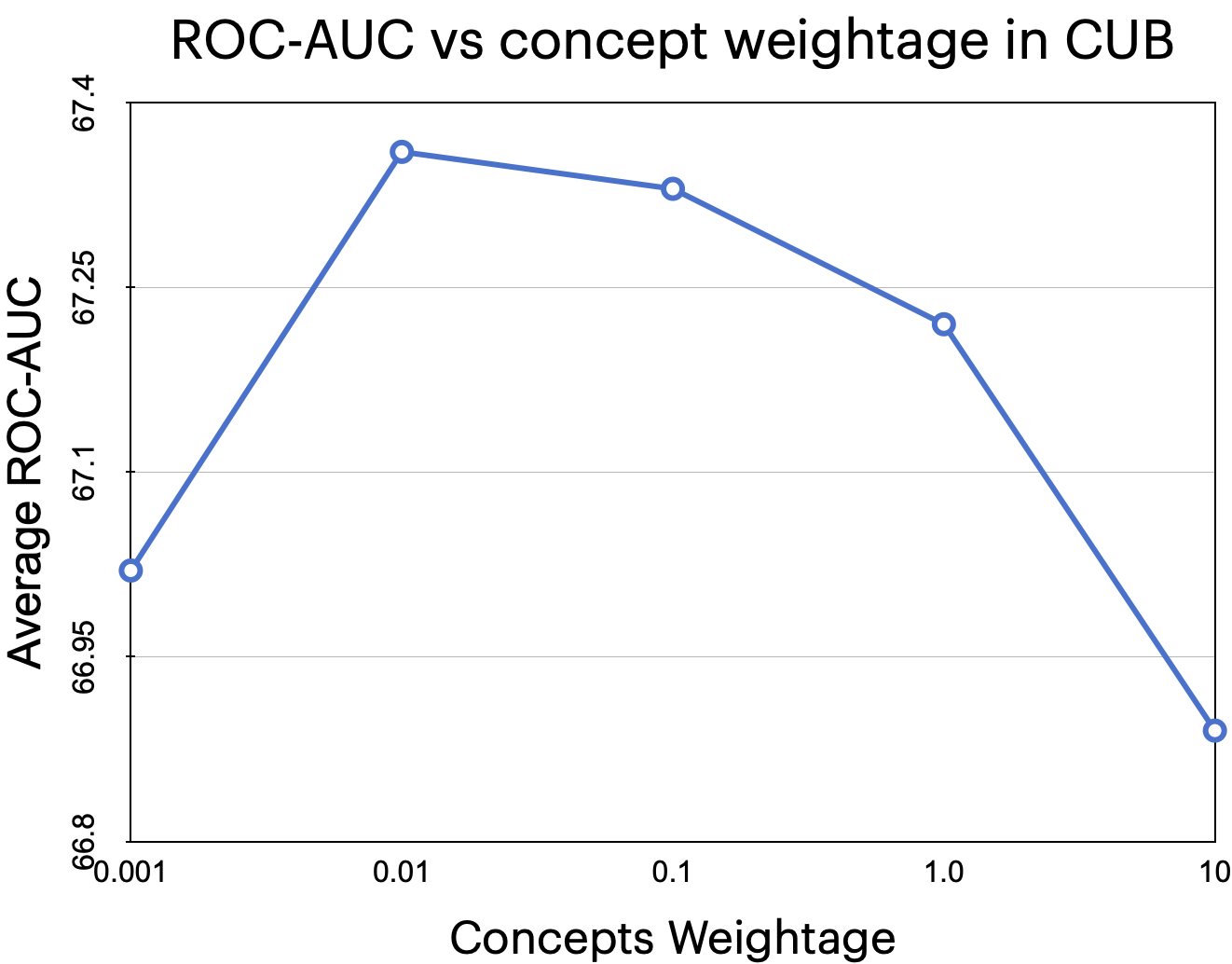} 
    \caption{The impact of concept weightage on CUB dataset: Higher values of $\alpha$ generally lead to a reduction in ROC-AUC.}
    \label{fig:weightagevscub}
\end{figure}

\begin{figure}[] 
    \centering
    \includegraphics[width=0.45 \textwidth]{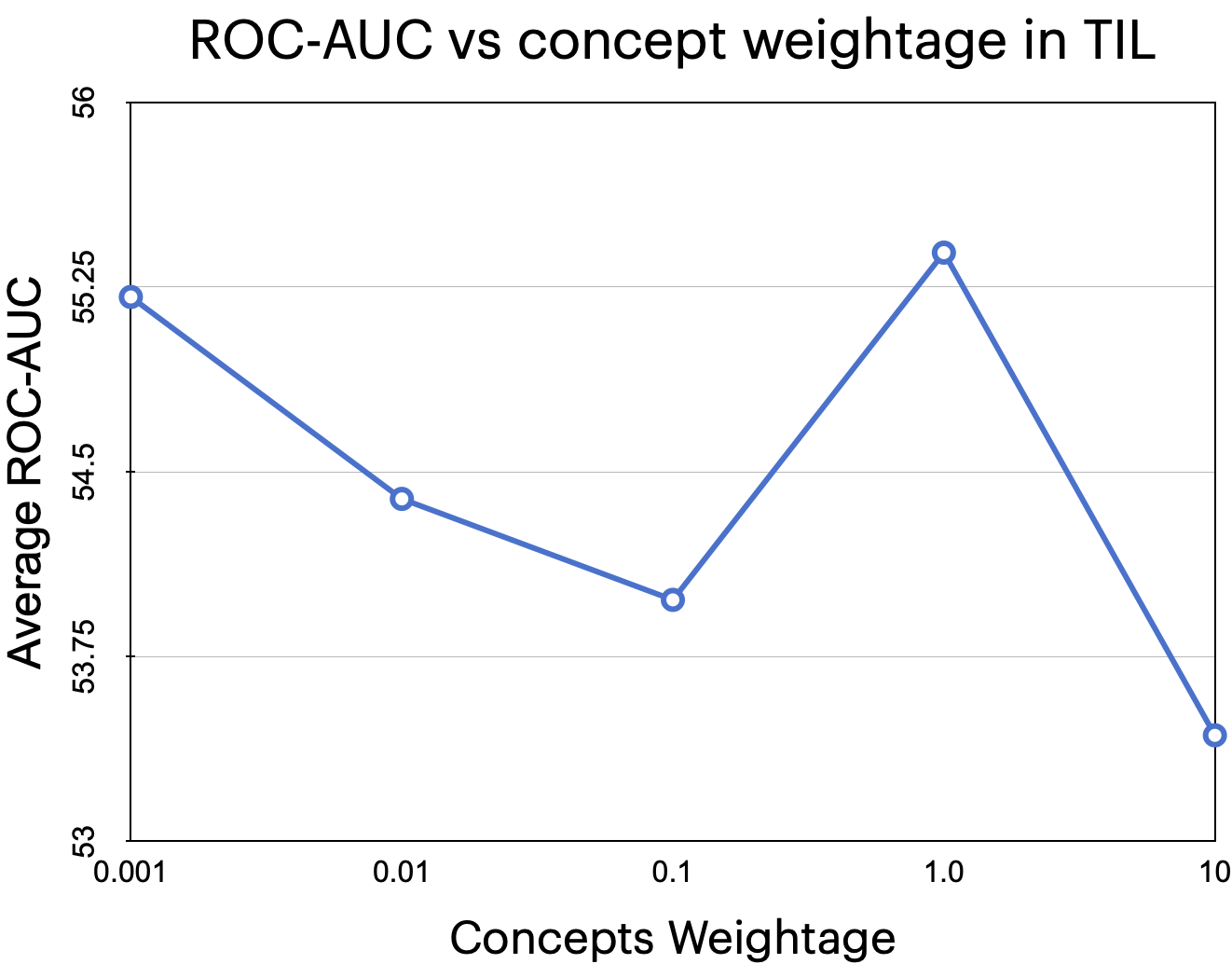} 
    \caption{The effect of concept weightage on TIL dataset; GOAD+ACE achieves highest ROC-AUC with $\alpha = 1.0$.}
    \label{fig:weightagevstil}
\end{figure}

\subsection{Extending ACE to other AD Methods}\label{sq:msc}
To prove that the explanatory feature of ACE is easily applicable to other anomaly detection models, we integrated a new AD model as the backbone for our anomaly detection framework. Therefore to evaluate the effectiveness of the concept explainability module introduced in ACE, we used a recent transformation-based anomaly detection model named MSC~\cite{reiss2023msc} and added the concept encoder to it.

For our experiments on MSC+ACE with a transformation model inspired by MSC~\cite{reiss2023msc}, we used similar hyperparameters. During the training, each sample within a batch will go over $M=2$ random transformations. We used pre-trained ResNet18 for our encoder $\mathcal{G}_{X}$. We trained MSC+ACE with SGD optimizer with learning rate $1  e-3$, attribute weight ($\alpha$) $0.1$. MSC+ACE is trained for 20 epochs for both dataset with batch size 8, 32 for CUB and TIL respectively. Similar hyperparameters are used for the experiments on MSC.

The results of MSC and MSC+ACE are presented in Table~\ref{tab:results_ab}. While the outcomes of our model, incorporating an anomaly detection backbone influenced by MSC~\cite{reiss2023msc}, reveal significantly enhanced performance compared to GOAD+ACE, this comes with certain considerations. MSC employs a k-Nearest Neighbor algorithm with $k=2$ to determine whether a sample is categorized as normal or anomalous, and this decision-making involves the storage of all training set embeddings. Consequently, there exists a trade-off between the desired accuracy and the available resources, particularly when dealing with larger datasets.

\begin{table}[t]
    \centering
    \begin{adjustbox}{width=0.5\textwidth}
    \begin{tabular}{lcccc}
    \toprule
    Datasets   & ~~ & CUB & TIL & CelebA \\
    \midrule
    MSC   & ~~ &\textbf{93.90}$\pm$2.90 &62.42$\pm$1.01 & 70.76$\pm$4.73 \\
    \midrule
    MSC+ACE     & ~~ &93.85$\pm$3.00 &\textbf{64.83}$\pm$1.32 & \textbf{70.99}$\pm$4.37 \\
    \bottomrule
    \end{tabular}
    \end{adjustbox}
    \renewcommand\thetable{2}
    \caption{\small{ROC-AUC ($\%$) comparison of MSC+ACE with MSC on \textit{one-vs-all} scheme on CUB and TIL datasets. All of the results are from our implementations and are averaged over five different runs for TIL and CUB and three for CelebA.}}
    \label{tab:results_ab}
\end{table}

\section{Limitations}
While ACE improves anomaly detection performance through human explainable concepts, its adaptability to diverse anomaly detection scenarios relies on the presence of annotated concepts. Recent works~\cite{oikarinen2023label,yuksekgonul2022post} overcame the limitation of concept annotations by querying a large vision-language model for concepts. However, as mentioned in \cite{yuksekgonul2022post}, the concepts are prone to model biases which are undesirable. Additionally, these models would fail to generate concepts in realistic datasets such as those in medical datasets. However to have a generalizable concept generating model is a fairly difficult task, although overcoming this bottleneck could be an interesting future work. Since our contribution is uncovering the black-box anomaly detection via human-interpretable concepts and is algorithmic in nature, we have not performed a user survey of using these concepts in anomaly detection. We hope that our work can motivate experiments in the medical domain.

\section{Conclusion and Future Work}

In this paper, we proposed a methodology that introduces transparency in anomaly detection prediction beyond standard metrics. While current anomaly detection approaches achieve promising performance, real-world applications demand transparency in model predictions beyond accuracy. Our proposal, transparent Anomaly detection Concept Explanations (ACE) addresses this challenge, offering human-interpretable insights along with anomaly prediction. Our experiments conducted on realistic datasets demonstrate comparable or better results in comparison to black box anomaly detectors. Additionally, we showcased the adaptability of the explanatory module to other transformation-based AD models. We hope that our work encourages research in interpreting and explaining anomaly detection. In future work, we intend to extend our experiments in exploring the integration of ACE into anomaly detection models beyond transformation-based detectors. Additionally, investigating the influence of intervention on anomaly detection performance is another avenue to explore.

\section{Acknowledgements}
The authors would like to acknowledge compute support by Digital Research Alliance of Canada. 
{\small
\bibliographystyle{ieee_fullname}
\bibliography{egbib}

\begin{thebibliography}{10}\itemsep=-1pt

\bibitem{adebayo2020debugging}
Julius Adebayo, Michael Muelly, Ilaria Liccardi, and Been Kim.
\newblock Debugging tests for model explanations.
\newblock {\em arXiv preprint arXiv:2011.05429}, 2020.

\bibitem{an2015variational}
Jinwon An and Sungzoon Cho.
\newblock Variational autoencoder based anomaly detection using reconstruction
  probability.
\newblock {\em Special Lecture on IE}, 2(1):1--18, 2015.

\bibitem{Bergman2020GOAD}
Liron Bergman and Yedid Hoshen.
\newblock Classification-based anomaly detection for general data.
\newblock In {\em International Conference on Learning Representations}, 2020.

\bibitem{carletti2023interpretable}
Mattia Carletti, Matteo Terzi, and Gian~Antonio Susto.
\newblock Interpretable anomaly detection with diffi: Depth-based feature
  importance of isolation forest.
\newblock {\em Engineering Applications of Artificial Intelligence},
  119:105730, 2023.

\bibitem{chandola2009anomaly}
Varun Chandola, Arindam Banerjee, and Vipin Kumar.
\newblock Anomaly detection: A survey.
\newblock {\em ACM computing surveys (CSUR)}, 41(3):1--58, 2009.

\bibitem{chauhan2023interactive}
Kushal Chauhan, Rishabh Tiwari, Jan Freyberg, Pradeep Shenoy, and Krishnamurthy
  Dvijotham.
\newblock Interactive concept bottleneck models.
\newblock In {\em Proceedings of the AAAI Conference on Artificial
  Intelligence}, volume~37, pages 5948--5955, 2023.

\bibitem{cho2023training}
Wonwoo Cho, Jeonghoon Park, and Jaegul Choo.
\newblock Training auxiliary prototypical classifiers for explainable anomaly
  detection in medical image segmentation.
\newblock In {\em Proceedings of the IEEE/CVF Winter Conference on Applications
  of Computer Vision}, pages 2624--2633, 2023.

\bibitem{deecke2018ADGAN}
Lucas Deecke, Robert Vandermeulen, Lukas Ruff, Stephan Mandt, and Marius Kloft.
\newblock Anomaly detection with generative adversarial networks, 2018.

\bibitem{doshi2023towards}
Keval Doshi and Yasin Yilmaz.
\newblock Towards interpretable video anomaly detection.
\newblock In {\em Proceedings of the IEEE/CVF Winter Conference on Applications
  of Computer Vision}, pages 2655--2664, 2023.

\bibitem{espinosa2022concept}
Mateo Espinosa~Zarlenga, Pietro Barbiero, Gabriele Ciravegna, Giuseppe Marra,
  Francesco Giannini, Michelangelo Diligenti, Zohreh Shams, Frederic Precioso,
  Stefano Melacci, Adrian Weller, et~al.
\newblock Concept embedding models: Beyond the accuracy-explainability
  trade-off.
\newblock {\em Advances in Neural Information Processing Systems},
  35:21400--21413, 2022.

\bibitem{ghosh2009snn}
Samanwoy Ghosh-Dastidar and Hojjat Adeli.
\newblock Third generation neural networks: Spiking neural networks.
\newblock In {\em Advances in computational intelligence}, pages 167--178.
  Springer, 2009.

\bibitem{Golan2018GEOM}
Izhak Golan and Ran El-Yaniv.
\newblock Deep anomaly detection using geometric transformations.
\newblock In S. Bengio, H. Wallach, H. Larochelle, K. Grauman, N. Cesa-Bianchi,
  and R. Garnett, editors, {\em Advances in Neural Information Processing
  Systems}, volume~31. Curran Associates, Inc., 2018.

\bibitem{havasi2022addressing}
Marton Havasi, Sonali Parbhoo, and Finale Doshi-Velez.
\newblock Addressing leakage in concept bottleneck models.
\newblock {\em Advances in Neural Information Processing Systems},
  35:23386--23397, 2022.

\bibitem{he2018triplet}
Xinwei He, Yang Zhou, Zhichao Zhou, Song Bai, and Xiang Bai.
\newblock Triplet-center loss for multi-view 3d object retrieval.
\newblock In {\em Proceedings of the IEEE conference on computer vision and
  pattern recognition}, pages 1945--1954, 2018.

\bibitem{hendrycks2018deep}
Dan Hendrycks, Mantas Mazeika, and Thomas Dietterich.
\newblock Deep anomaly detection with outlier exposure.
\newblock In {\em International Conference on Learning Representations}, 2019.

\bibitem{jianliang2009k_meansAD}
Meng Jianliang, Shang Haikun, and Bian Ling.
\newblock The application on intrusion detection based on k-means cluster
  algorithm.
\newblock In {\em 2009 International Forum on Information Technology and
  Applications}, volume~1, pages 150--152. IEEE, 2009.

\bibitem{kim2018interpretability}
Been Kim, Martin Wattenberg, Justin Gilmer, Carrie Cai, James Wexler, Fernanda
  Viegas, et~al.
\newblock Interpretability beyond feature attribution: Quantitative testing
  with concept activation vectors (tcav).
\newblock In {\em International conference on machine learning}, pages
  2668--2677. PMLR, 2018.

\bibitem{kim2023probabilistic}
Eunji Kim, Dahuin Jung, Sangha Park, Siwon Kim, and Sungroh Yoon.
\newblock Probabilistic concept bottleneck models.
\newblock {\em arXiv preprint arXiv:2306.01574}, 2023.

\bibitem{kirillov2023sam}
Alexander Kirillov, Eric Mintun, Nikhila Ravi, Hanzi Mao, Chloe Rolland, Laura
  Gustafson, Tete Xiao, Spencer Whitehead, Alexander~C Berg, Wan-Yen Lo, et~al.
\newblock Segment anything.
\newblock {\em arXiv preprint arXiv:2304.02643}, 2023.

\bibitem{koh2020cbm}
Pang~Wei Koh, Thao Nguyen, Yew~Siang Tang, Stephen Mussmann, Emma Pierson, Been
  Kim, and Percy Liang.
\newblock Concept bottleneck models.
\newblock In {\em International conference on machine learning}, pages
  5338--5348. PMLR, 2020.

\bibitem{latecki2007KDE}
Longin~Jan Latecki, Aleksandar Lazarevic, and Dragoljub Pokrajac.
\newblock Outlier detection with kernel density functions.
\newblock In {\em International Workshop on Machine Learning and Data Mining in
  Pattern Recognition}, pages 61--75. Springer, 2007.

\bibitem{lewis2020bart}
Mike Lewis, Yinhan Liu, Naman Goyal, Marjan Ghazvininejad, Abdelrahman Mohamed,
  Omer Levy, Veselin Stoyanov, and Luke Zettlemoyer.
\newblock {BART}: Denoising sequence-to-sequence pre-training for natural
  language generation, translation, and comprehension.
\newblock In {\em Proceedings of the 58th Annual Meeting of the Association for
  Computational Linguistics}, pages 7871--7880, Online, July 2020. Association
  for Computational Linguistics.

\bibitem{liu2018large}
Ziwei Liu, Ping Luo, Xiaogang Wang, and Xiaoou Tang.
\newblock Large-scale celebfaces attributes (celeba) dataset.
\newblock {\em Retrieved August}, 15(2018):11, 2018.

\bibitem{mahinpei2021promises}
Anita Mahinpei, Justin Clark, Isaac Lage, Finale Doshi-Velez, and Weiwei Pan.
\newblock Promises and pitfalls of black-box concept learning models.
\newblock {\em arXiv preprint arXiv:2106.13314}, 2021.

\bibitem{Mathieu2016video_fram_prediction}
Michael Mathieu, Camille Couprie, and Yann LeCun.
\newblock Deep multi-scale video prediction beyond mean square error.
\newblock In {\em International Conference on Learning Representations}, 2016.

\bibitem{nichol2021glide}
Alex Nichol, Prafulla Dhariwal, Aditya Ramesh, Pranav Shyam, Pamela Mishkin,
  Bob McGrew, Ilya Sutskever, and Mark Chen.
\newblock Glide: Towards photorealistic image generation and editing with
  text-guided diffusion models.
\newblock {\em arXiv preprint arXiv:2112.10741}, 2021.

\bibitem{oikarinen2023label}
Tuomas Oikarinen, Subhro Das, Lam~M Nguyen, and Tsui-Wei Weng.
\newblock Label-free concept bottleneck models.
\newblock {\em arXiv preprint arXiv:2304.06129}, 2023.

\bibitem{parzen1962estimation}
Emanuel Parzen.
\newblock On estimation of a probability density function and mode.
\newblock {\em The annals of mathematical statistics}, 33(3):1065--1076, 1962.

\bibitem{raffel2020t5}
Colin Raffel, Noam Shazeer, Adam Roberts, Katherine Lee, Sharan Narang, Michael
  Matena, Yanqi Zhou, Wei Li, and Peter~J Liu.
\newblock Exploring the limits of transfer learning with a unified text-to-text
  transformer.
\newblock {\em The Journal of Machine Learning Research}, 21(1):5485--5551,
  2020.

\bibitem{rafiee2020unsupervised}
Laya Rafiee and Thomas Fevens.
\newblock Unsupervised anomaly detection with a gan augmented autoencoder.
\newblock In {\em International Conference on Artificial Neural Networks},
  pages 479--490. Springer, 2020.

\bibitem{reiss2023msc}
Tal Reiss and Yedid Hoshen.
\newblock Mean-shifted contrastive loss for anomaly detection.
\newblock In {\em Proceedings of the AAAI Conference on Artificial
  Intelligence}, volume~37, pages 2155--2162, 2023.

\bibitem{ruff2018deep_SVDD}
Lukas Ruff, Robert Vandermeulen, Nico Goernitz, Lucas Deecke, Shoaib~Ahmed
  Siddiqui, Alexander Binder, Emmanuel M{\"u}ller, and Marius Kloft.
\newblock Deep one-class classification.
\newblock In {\em International conference on machine learning}, pages
  4393--4402. PMLR, 2018.

\bibitem{Saltz2018SpatialOA}
J. Saltz, Rajarsi~R. Gupta, Le Hou, Tahsin~M. Kurç, Pankaj~Kumar Singh, Vu
  Nguyen, Dimitris Samaras, Kenneth~R Shroyer, Tianhao Zhao, Rebecca~C.
  Batiste, John S.~Van Arnam, Ilya Shmulevich, Arvind U.~K. Rao, Alexander~J.
  Lazar, Ashish Sharma, and V{\'e}steinn Thorsson.
\newblock Spatial organization and molecular correlation of tumor-infiltrating
  lymphocytes using deep learning on pathology images.
\newblock {\em Cell reports}, 23 1:181--193.e7, 2018.

\bibitem{schlegl2017AnoGAN}
Thomas Schlegl, Philipp Seeb{\"o}ck, Sebastian~M Waldstein, Ursula
  Schmidt-Erfurth, and Georg Langs.
\newblock Unsupervised anomaly detection with generative adversarial networks
  to guide marker discovery.
\newblock In {\em Proc.~of IPMI}, pages 146--157. Springer, 2017.

\bibitem{scholkopf1999OC_SVM}
Bernhard Sch{\"o}lkopf, Robert~C Williamson, Alexander~J Smola, John
  Shawe-Taylor, John~C Platt, et~al.
\newblock Support vector method for novelty detection.
\newblock In {\em NIPS}, volume~12, pages 582--588. Citeseer, 1999.

\bibitem{seras2022novel}
Aitor~Martinez Seras, Javier Del~Ser, Jesus~L Lobo, Pablo Garcia-Bringas, and
  Nikola Kasabov.
\newblock A novel explainable out-of-distribution detection approach for
  spiking neural networks.
\newblock {\em arXiv preprint arXiv:2210.00894}, 2022.

\bibitem{sevyeri2022adcgan}
Laya~Rafiee Sevyeri and Thomas Fevens.
\newblock Ad-cgan: Contrastive generative adversarial network for anomaly
  detection.
\newblock In {\em International Conference on Image Analysis and Processing},
  pages 322--334. Springer, 2022.

\bibitem{sheth2023overcoming}
Ivaxi Sheth and Samira~Ebrahimi Kahou.
\newblock Overcoming interpretability and accuracy trade-off in medical
  imaging.
\newblock In {\em Medical Imaging with Deep Learning, short paper track}, 2023.

\bibitem{sheth2022learning}
Ivaxi Sheth, Aamer~Abdul Rahman, Laya~Rafiee Sevyeri, Mohammad Havaei, and
  Samira~Ebrahimi Kahou.
\newblock Learning from uncertain concepts via test time interventions.
\newblock In {\em Workshop on Trustworthy and Socially Responsible Machine
  Learning, NeurIPS 2022}, 2022.

\bibitem{shin2022a}
Sungbin Shin, Yohan Jo, Sungsoo Ahn, and Namhoon Lee.
\newblock A closer look at the intervention procedure of concept bottleneck
  models.
\newblock In {\em Workshop on Trustworthy and Socially Responsible Machine
  Learning, NeurIPS 2022}, 2022.

\bibitem{sohn2021learning}
Kihyuk Sohn, Chun-Liang Li, Jinsung Yoon, Minho Jin, and Tomas Pfister.
\newblock Learning and evaluating representations for deep one-class
  classification.
\newblock In {\em International Conference on Learning Representations}, 2021.

\bibitem{sundararajan2017axiomatic}
Mukund Sundararajan, Ankur Taly, and Qiqi Yan.
\newblock Axiomatic attribution for deep networks.
\newblock In {\em International conference on machine learning}, pages
  3319--3328. PMLR, 2017.

\bibitem{szymanowicz2022discrete}
Stanislaw Szymanowicz, James Charles, and Roberto Cipolla.
\newblock Discrete neural representations for explainable anomaly detection.
\newblock In {\em Proceedings of the IEEE/CVF winter conference on applications
  of computer vision}, pages 148--156, 2022.

\bibitem{tack2020csi}
Jihoon Tack, Sangwoo Mo, Jongheon Jeong, and Jinwoo Shin.
\newblock Csi: Novelty detection via contrastive learning on distributionally
  shifted instances.
\newblock {\em Advances in neural information processing systems},
  33:11839--11852, 2020.

\bibitem{tax2004SVDD}
David~MJ Tax and Robert~PW Duin.
\newblock Support vector data description.
\newblock {\em Machine learning}, 54(1):45--66, 2004.

\bibitem{Welinder2010CaltechUCSDB2}
Peter Welinder, Steve Branson, Takeshi Mita, Catherine Wah, Florian Schroff,
  Serge~J. Belongie, and Pietro Perona.
\newblock Caltech-ucsd birds 200.
\newblock 2010.

\bibitem{xu2023interpretable}
Romain Xu-Darme, Julien Girard-Satabin, Darryl Hond, Gabriele Incorvaia, and
  Zakaria Chihani.
\newblock Interpretable out-of-distribution detection using pattern
  identification.
\newblock {\em arXiv preprint arXiv:2302.10303}, 2023.

\bibitem{xu2022particul}
Romain Xu-Darme, Georges Qu{\'e}not, Zakaria Chihani, and Marie-Christine
  Rousset.
\newblock Particul: Part identification with confidence measure using
  unsupervised learning.
\newblock {\em arXiv preprint arXiv:2206.13304}, 2022.

\bibitem{yang2017towards}
Bo Yang, Xiao Fu, Nicholas~D Sidiropoulos, and Mingyi Hong.
\newblock Towards k-means-friendly spaces: Simultaneous deep learning and
  clustering.
\newblock In {\em international conference on machine learning}, pages
  3861--3870. PMLR, 2017.

\bibitem{yang2021insloc}
Ceyuan Yang, Zhirong Wu, Bolei Zhou, and Stephen Lin.
\newblock Instance localization for self-supervised detection pretraining.
\newblock In {\em CVPR}, 2021.

\bibitem{yuksekgonul2022post}
Mert Yuksekgonul, Maggie Wang, and James Zou.
\newblock Post-hoc concept bottleneck models.
\newblock {\em arXiv preprint arXiv:2205.15480}, 2022.

\bibitem{zagoruyko2016wide}
Sergey Zagoruyko and Nikos Komodakis.
\newblock Wide residual networks.
\newblock {\em arXiv preprint arXiv:1605.07146}, 2016.

\bibitem{zarlenga2023learning}
Mateo~Espinosa Zarlenga, Katherine~M Collins, Krishnamurthy Dvijotham, Adrian
  Weller, Zohreh Shams, and Mateja Jamnik.
\newblock Learning to receive help: Intervention-aware concept embedding
  models.
\newblock {\em arXiv preprint arXiv:2309.16928}, 2023.

\bibitem{zenati2018efficient}
Houssam Zenati, Chuan~Sheng Foo, Bruno Lecouat, Gaurav Manek, and
  Vijay~Ramaseshan Chandrasekhar.
\newblock Efficient gan-based anomaly detection.
\newblock {\em arXiv preprint arXiv:1802.06222}, 2018.

\bibitem{zenati2018ALAD}
Houssam Zenati, Manon Romain, Chuan-Sheng Foo, Bruno Lecouat, and Vijay
  Chandrasekhar.
\newblock Adversarially learned anomaly detection.
\newblock In {\em 2018 IEEE International Conference on Data Mining (ICDM)},
  pages 727--736. IEEE, 2018.

\bibitem{zhang2016colorful}
Richard Zhang, Phillip Isola, and Alexei~A Efros.
\newblock Colorful image colorization.
\newblock In {\em European conference on computer vision}, pages 649--666.
  Springer, 2016.

\bibitem{zhou2017anomaly}
Chong Zhou and Randy~C Paffenroth.
\newblock Anomaly detection with robust deep autoencoders.
\newblock In {\em Proceedings of the 23rd ACM SIGKDD International Conference
  on Knowledge Discovery and Data Mining}, pages 665--674. ACM, 2017.

\bibitem{zong2018gaussian_mixture}
Bo Zong, Qi Song, Martin~Renqiang Min, Wei Cheng, Cristian Lumezanu, Daeki Cho,
  and Haifeng Chen.
\newblock Deep autoencoding gaussian mixture model for unsupervised anomaly
  detection.
\newblock In {\em International Conference on Learning Representations}, 2018.

\end{thebibliography}
}

\end{document}